\pdfoutput=1
\documentclass[10pt,twocolumn]{article}

\usepackage[letterpaper,margin=0.75in]{geometry}
\usepackage{microtype}
\usepackage{amsmath,amsfonts}
\usepackage{amssymb}
\usepackage{mathtools}
\usepackage{array}
\usepackage{booktabs}
\usepackage[table]{xcolor}
\usepackage{graphicx}
\usepackage{textcomp}
\usepackage{url}
\usepackage{cite}
\usepackage{pifont}
\usepackage{caption}

\definecolor{oursrow}{RGB}{232,244,252}

\usepackage[colorlinks=true,linkcolor=blue,citecolor=blue,urlcolor=blue]{hyperref}
\hypersetup{
    pdftitle={Reasoning as Intersection: Consensus-Frame Alignment for Visual Focus in Video-MLLMs},
    pdfauthor={Chengwen Liu, Zhe Huang, Jisheng Dang, Hong Peng, Qi Tian, Tat-Seng Chua}
}

\title{Reasoning as Intersection: Consensus-Frame Alignment for Visual Focus in Video-MLLMs}

\author{
Chengwen Liu$^{1,*}$ \quad
Zhe Huang$^{2,*}$ \quad
Jisheng Dang$^{1}$ \quad
Hong Peng$^{1,\dagger}$ \quad
Qi Tian$^{3}$ \quad
Tat-Seng Chua$^{4}$\\[0.5em]
\small $^1$School of Information Science and Engineering, Lanzhou University, Lanzhou, China\\
\small $^2$Beijing University of Posts and Telecommunications, Beijing, China\\
\small $^3$Cloud and AI BU, Huawei, Shenzhen, China\\
\small $^4$School of Computing, National University of Singapore, Singapore\\[0.5em]
\small $^*$Equal contribution. \quad $^\dagger$Corresponding author: Hong Peng.
}

\date{}

\begin{document}

\maketitle

\def\method{CF-GRPO}

\begin{abstract}
Reinforcement learning has improved the reasoning ability of large language models, but applying outcome-only rewards to video multimodal large language models (Video-MLLMs) provides limited guidance on which visual evidence should support the answer. Inspired by multisensory integration, where consistent cues can enhance the salience and reliability of perceptual estimates, we introduce Consensus Frame GRPO (CF-GRPO), a temporal-annotation-free process-level reward framework for evidence-aware video reasoning. CF-GRPO constructs a consensus frame prior from intrinsic video cues, including temporal coverage, scene-transition cues, and query-conditioned visual relevance. It then computes a model-side frame-use score from visual and response representations and optimizes their agreement through the Consensus Frame Reward (CFR). With salience-aware sparse aggregation and distribution sharpening, CFR provides a high-contrast reward signal without requiring human temporal annotations. Experiments show that VideoCFR achieves competitive performance across complex video reasoning benchmarks and improves several metrics over representative Video-MLLM and RL baselines, while the consensus prior provides an interpretable view of the evidence frames emphasized during training. The implementation is available at \url{https://github.com/1Pansy/VideoCFR}.
\end{abstract}

\noindent\textbf{Keywords:}
Video-MLLMs, video reasoning, reinforcement learning, process-level reward, frame evidence alignment.

\begin{figure*}[!t]
  \begin{center}
    \centerline{\includegraphics[width=\textwidth]{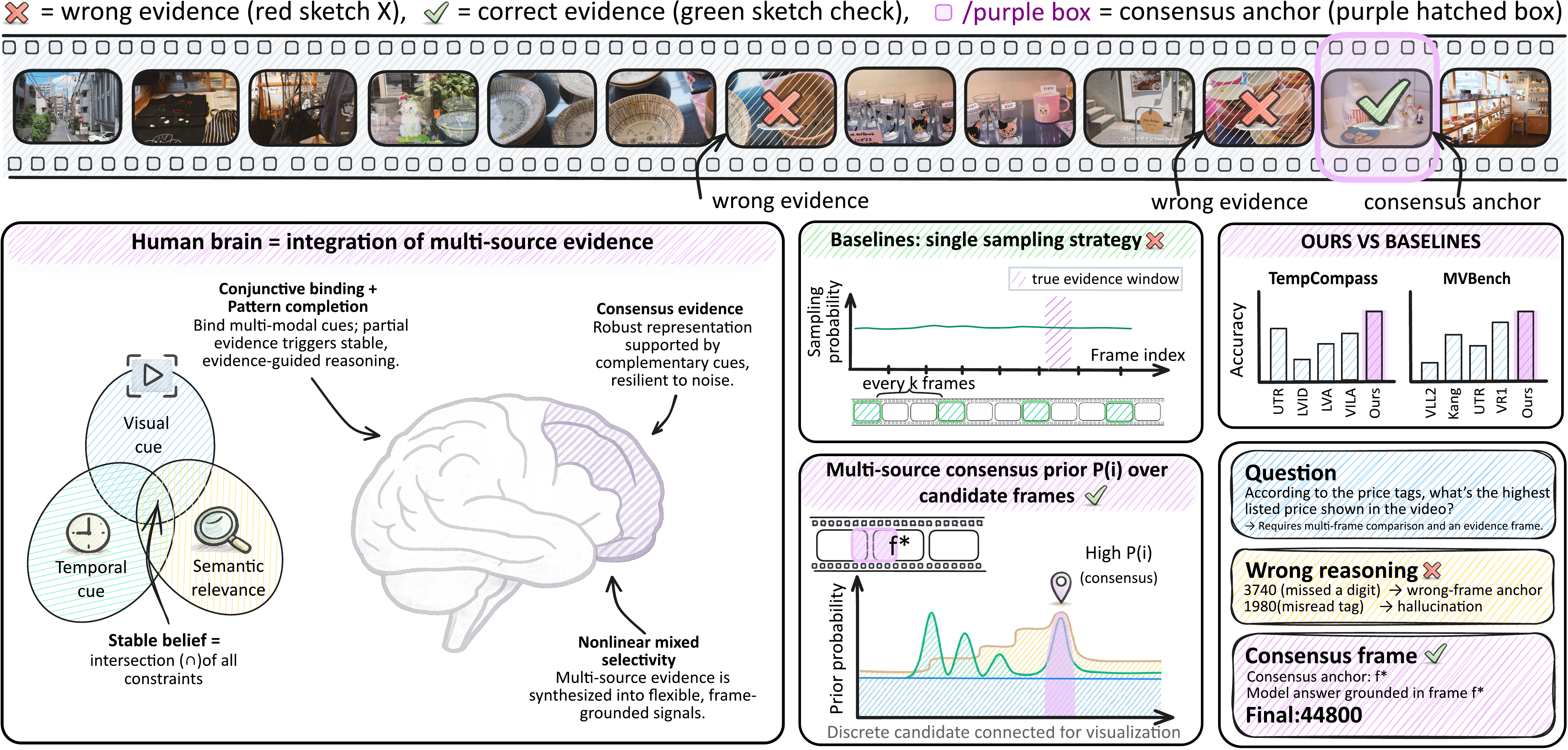}}
    \caption{Motivation of consensus-frame alignment. In long-video QA, single-source sampling or diffuse frame use can anchor the response to visually plausible but incorrect frames. VideoCFR constructs a consensus prior from temporal coverage, scene-transition, and query-relevance cues, and uses this prior as training-time evidence guidance for aligning generation with frames that support the answer.}
    \label{fig:intro}
  \end{center}
\end{figure*}

\section{Introduction}

Recent advances in reinforcement learning (RL), including group relative policy optimization (GRPO), have made post-training an effective way to align large language models (LLMs) and improve their reasoning abilities \cite{ouyang2022training, shao2024deepseekmath}. This direction has also been adapted to video multimodal large language models (Video-MLLMs), where recent methods use answer correctness, temporal consistency, localization, tool-use, or task-specific verifiable rewards to improve video reasoning and grounding \cite{videor1, park2025deepvideo, li2025videochat, VIDEORFT, Temporal-RLT, wang2026versatilevideo, zhang2026thinkingvideos}. These studies show that RL can improve Video-MLLM behavior when the reward is better matched to video tasks.

Despite this progress, video reasoning remains under-specified when supervision is dominated by final-answer correctness. A video contains redundant, temporally distributed, and heterogeneous visual evidence. A rollout-level answer reward can indicate whether the response is correct, but it does not specify which frames should support the answer, whether the model has relied on visually relevant evidence, or whether a visually salient but irrelevant frame has dominated the response. Temporal rewards and localization rewards address important aspects of video understanding, but they usually supervise temporal order, answer validity, or explicit grounding targets rather than the frame-level evidence used during generation. Therefore, Video-MLLM RL needs a process signal that is closer to the visual evidence used by the model while still preserving standard outcome supervision.

In \autoref{fig:intro}, we illustrate this motivation. In long or untrimmed videos, a single sampling strategy or a diffuse frame-use pattern can anchor the response to frames that are plausible but not decisive for the question. In the example, the model must identify the evidence frame containing the relevant price tag and compare it with the question requirement; attending to another visually similar frame leads to an incorrect answer. This failure mode is not simply a lack of more frames. It is a mismatch between the evidence needed by the question and the frames emphasized by the generation process.

Existing evidence-selection and video-grounding methods also recognize the importance of informative frames, using keyframe selection, token compression, semantic-visual consensus, or spatio-temporal grounding to reduce redundancy or localize events \cite{tang2025adaptive, zhu2025focus, sheng2025sevices, qin2026efficientframe, wu2026marc, wang2025spacevllm}. These methods are closely related, but our goal is different. We do not aim to choose a final input subset at inference time or add a grounding head with temporal annotations. Instead, we ask whether RL training itself can receive a frame-level signal that encourages the generated response to be associated with candidate evidence frames.

To obtain such a signal without manual temporal labels, our design uses a consensus prior over frames. The motivation is consistent with multisensory integration in biological perception, where information from multiple channels can improve event salience and estimate reliability under uncertainty \cite{stein2008multisensory, ernst2002humans, alais2004ventriloquist, knill2004bayesian}. For video reasoning, we instantiate this idea through intrinsic video cues rather than through human annotations: uniform temporal coverage preserves the global sequence, scene transitions capture visual changes, and query-conditioned visual relevance identifies frames related to the question. Their agreement defines a soft consensus frame prior. This prior is not treated as ground-truth temporal annotation; it is a temporal-annotation-free estimate of which candidate frames are more likely to contain useful evidence.

We introduce Consensus Frame GRPO (CF-GRPO), which supplements standard answer rewards with the Consensus Frame Reward (CFR). CFR encourages agreement between the consensus frame prior and a model-side frame-use score extracted from visual and response representations. This design is also related to process supervision, which provides feedback on intermediate reasoning behavior rather than only final answers \cite{uesato2022solving, lightman2023lets, wang2025visualprm, zhang2026pearl}. However, video evidence alignment requires feedback over temporally distributed visual frames, whereas existing process rewards mainly evaluate textual reasoning steps or image-conditioned reasoning traces. CFR therefore converts process supervision into a frame-level reward for video RL without requiring human temporal annotations.

CF-GRPO changes the training objective rather than only filtering frames before inference. During training, the consensus prior is compared with the model-side frame-use distribution, and their overlap provides a scalar reward that can be optimized together with accuracy, structural, and temporal rewards. To make this reward more discriminative, CFR uses salience-aware sparse aggregation to preserve high-response visual regions and temperature sharpening to reduce overly diffuse frame scores. Experiments show that VideoCFR obtains competitive performance on complex video reasoning benchmarks, and ablations indicate that the consensus prior, sparse aggregation, and sharpening each contribute to the final performance. Visual analyses further show that high-attention response-frame events are more concentrated on consensus frames, which is consistent with the intended evidence-alignment effect.

Our main contributions are as follows:

\begin{itemize}
\item 
We propose Consensus Frame GRPO, a temporal-annotation-free process-level RL framework for video reasoning. It augments outcome rewards with frame-level evidence alignment while avoiding expensive human temporal annotations.
\item 
We develop the Consensus Frame Reward mechanism. It constructs a consensus prior from temporal coverage, scene-transition, and query-conditioned visual relevance, then aligns this prior with a model-side frame-use score through salience-aware sparse aggregation and distribution sharpening.
\item
Our VideoCFR model achieves competitive performance on complex video reasoning benchmarks, and ablation and visualization analyses show that the proposed reward provides useful evidence-level training signals and interpretable frame-level priors.
\end{itemize}

\section{Related Work}
\begin{figure*}[!t]
  \begin{center}
    \centerline{\includegraphics[width=0.95\textwidth]{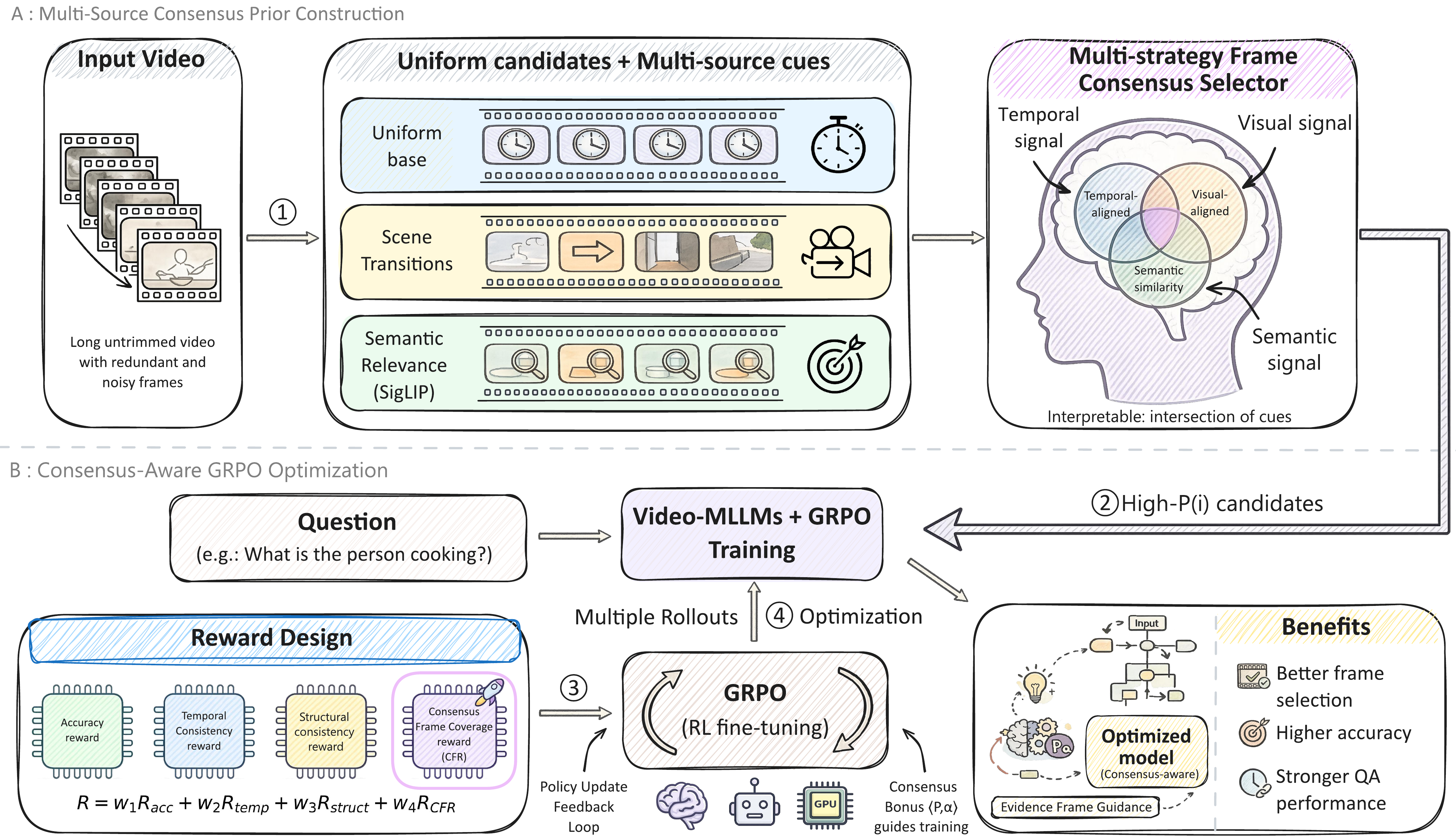}}
    \caption{Overview of the CF-GRPO framework. Panel A constructs a multi-source consensus prior from uniform coverage, scene transitions, and query-conditioned semantic relevance. Panel B incorporates CFR into GRPO, rewarding overlap between the consensus prior and the model-side frame-use distribution while preserving accuracy, temporal, and structural rewards.}
    \label{fig:main}
  \end{center}
\end{figure*}

\subsection{Reinforcement Learning for Video-MLLMs}

RL has become an effective post-training technique for aligning model behavior and improving reasoning in language models and multimodal models \cite{ouyang2022training, shao2024deepseekmath}. In the video domain, recent work adapts GRPO-style optimization to Video-MLLMs by defining verifiable rewards for video QA, temporal ordering, and grounding. Video-R1 introduces T-GRPO, which combines rule-based answer rewards with an ordered-versus-shuffled temporal reward to reduce reliance on single-frame shortcuts \cite{videor1}. DeepVideo-R1 studies optimization instability in video reinforcement fine-tuning and introduces Regressive GRPO with difficulty-aware data augmentation \cite{park2025deepvideo}. VideoChat-R1 applies reinforcement fine-tuning to spatio-temporal perception tasks, including temporal grounding and object tracking, showing that task-specific reward design can improve grounded video QA \cite{li2025videochat}. Temporal-RLT further analyzes reward design and data efficiency for VideoLLM reinforcement learning \cite{Temporal-RLT}. Recent studies also investigate data-efficient video RL and tool-augmented long-video reasoning, indicating that video-specific reward design is becoming a central issue in Video-MLLM post-training \cite{wang2026versatilevideo, zhang2026thinkingvideos}.

Several related studies broaden this direction by improving training recipes, model scales, or temporal reward construction. TinyLLaVA-Video-R1 investigates whether smaller multimodal models can acquire video reasoning ability through reinforcement learning \cite{TinyLLaVA-Video-R1}. VideoRFT focuses on reinforced fine-tuning for video reasoning capability \cite{VIDEORFT}. VIDEO-UTR studies temporal reward construction for scalable Video-MLLM training \cite{VIDEO-UTR}. These methods are closely related to ours because they share the same post-training setting and aim to make reward signals more suitable for video. However, their supervision is mainly expressed through final answer correctness, temporal consistency, or task-specific grounding labels. Such rewards indicate whether a response is correct or temporally consistent, but they provide limited direct guidance on which visual evidence frames should support the generated answer. CFR addresses this missing signal by introducing a frame-level evidence-alignment reward while retaining standard outcome and temporal rewards.

\subsection{Process Supervision and Reward Modeling}

Process supervision provides feedback on intermediate reasoning states rather than only on final answers. In text reasoning, process- and outcome-based supervision have been compared on mathematical problem solving, and step-level reward models have been shown to provide informative training and selection signals \cite{uesato2022solving, lightman2023lets}. This line of work motivates the use of denser feedback when final-answer supervision is under-specified. For multimodal reasoning, VisualPRM extends process reward modeling to visual tasks by training a multimodal reward model and evaluating step-wise correctness in reasoning traces \cite{wang2025visualprm}. Perceptual-evidence anchored RL further emphasizes that multimodal reasoning rewards can benefit from explicitly connecting reasoning behavior with perceptual evidence \cite{zhang2026pearl}.

Despite their relevance, existing process-supervision methods do not directly solve the video evidence-alignment problem considered here. Textual process rewards usually supervise logical steps, while multimodal PRMs typically evaluate reasoning traces or image-conditioned steps. Video reasoning additionally requires identifying temporally distributed evidence from redundant frame sequences. Directly collecting human labels for which frames support each answer would be expensive and dataset-dependent. Our method therefore constructs process feedback from intrinsic video cues. The consensus prior is not treated as ground-truth temporal annotation; instead, it defines a temporal-annotation-free frame-level prior that is used to regularize the model-side frame-use score during RL.

\subsection{Video Evidence Selection and Spatio-Temporal Grounding}

\begin{figure*}[!t]
  \begin{center}
    \centerline{\includegraphics[width=\textwidth]{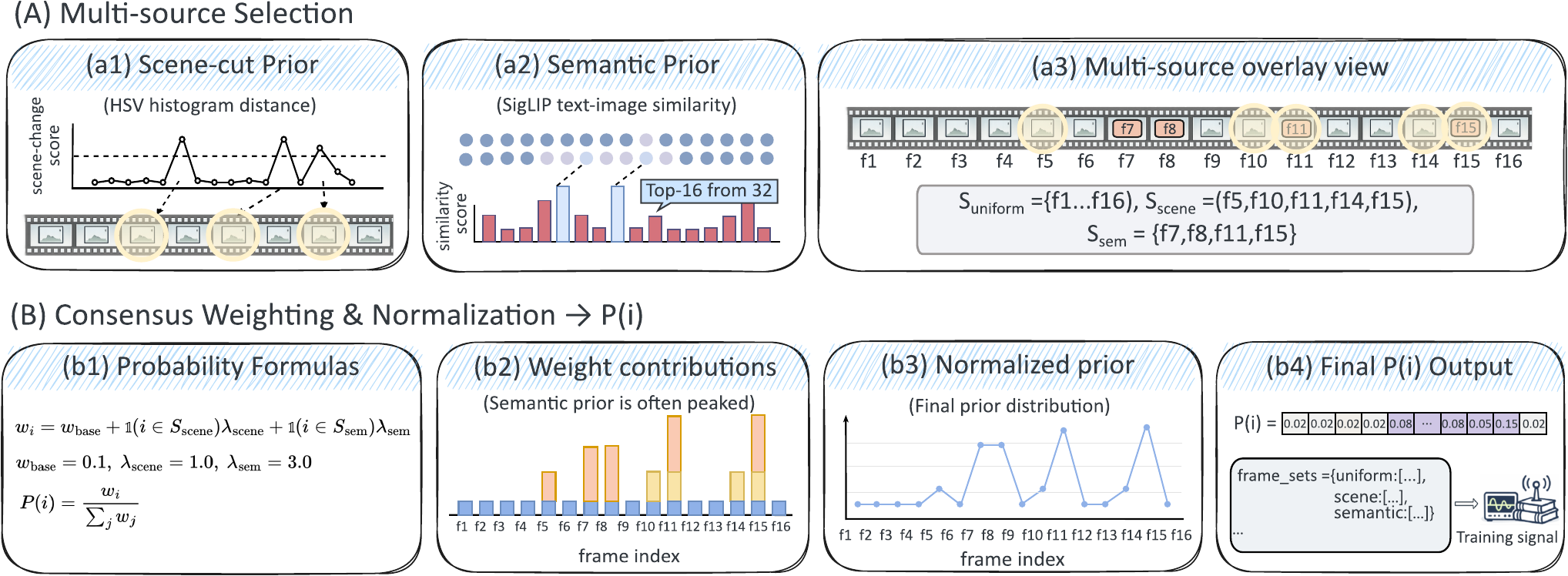}}
    \caption{Multi-source consensus prior construction. Candidate frames are characterized through uniform coverage, scene-cut responses, and query-conditioned semantic relevance. These cues are fused through weighted scoring and normalization to obtain $P(i)$, a soft temporal-annotation-free prior for downstream evidence alignment rather than a ground-truth temporal label.}
    \label{fig:sample}
  \end{center}
\end{figure*}

Another closely related direction addresses video redundancy by selecting, compressing, or explicitly grounding visual evidence before generation. Video-MLLMs have explored compact frame representations, long-context transfer, and video-oriented training recipes to process richer temporal context \cite{Llama-vid, VIDEOLLAMA2, LONGVA, LLAVA-ONEVISION, KANGEROO}. These approaches improve the capacity to ingest video context, but they do not by themselves specify which frames should be emphasized during RL optimization.

Keyframe selection methods make this evidence problem explicit. Adaptive Keyframe Sampling formulates long-video understanding as an input-side frame selection problem under a limited frame budget \cite{tang2025adaptive}. FOCUS similarly studies efficient keyframe selection for long-video understanding \cite{zhu2025focus}. SeViCES uses semantic-visual evidence consensus to identify informative frames and refine answers without retraining the backbone model \cite{sheng2025sevices}. Efficient Frame Selection further trains a selector with RL so that retained frames improve downstream video understanding \cite{qin2026efficientframe}. These methods show that relevance, coverage, and semantic-visual agreement are useful criteria for reducing video redundancy.

Spatio-temporal grounding methods pursue a related but different goal: they aim to localize when and where queried events or objects occur. SpaceVLLM introduces spatio-temporal aware queries and a query-guided space decoder, together with grounding data, to improve explicit localization in videos \cite{wang2025spacevllm}. Such methods provide strong grounding capabilities, but they usually require architectural components, grounding annotations, or task-specific localization objectives.

CFR is complementary to both evidence selection and grounding. It does not select a final input subset at inference time, and it does not add a grounding head to the Video-MLLM. Instead, it converts multiple evidence cues, including temporal coverage, scene transitions, and query-conditioned visual relevance, into a consensus prior for RL training. The reward then encourages agreement between this prior and the model-side frame-use score extracted from visual and response representations. Therefore, our contribution lies in process-level reward design for evidence alignment, rather than input compression, training-free keyframe selection, or architecture-level grounding.

\section{Method}

\subsection{Overview of Consensus Frame GRPO}
As illustrated in \autoref{fig:main}, CF-GRPO augments standard Video-MLLM RL with a temporal-annotation-free process reward. The goal is not to replace answer correctness, but to reduce reward underspecification by adding a frame-level signal about whether the generated response is aligned with candidate visual evidence.

In Panel A, we construct a consensus prior by fusing multiple complementary cues, including uniform temporal coverage, scene transitions, and query-conditioned visual relevance. This prior highlights frames that contain informative evidence while retaining broad temporal coverage.

In Panel B, the consensus prior is used to compute an auxiliary reward during Video-MLLM training. A model-side frame-use score is extracted from the similarity between visual frame representations and response hidden states. The Consensus Frame Reward (CFR) rewards agreement between this score and the consensus prior, alongside standard accuracy, structure, and temporal consistency rewards. This optimization provides evidence-level feedback without requiring human temporal labels.

\paragraph{Preliminaries} We consider an original video sequence with $T$ frames and uniformly sample $K$ frames as visual input. Unless otherwise noted, the frame index $i$ refers to one of these $K$ sampled frames, and all frame-level distributions are defined over the sampled frame set. The model generates responses conditioned on a query $q$ and the sampled video frames. Training employs GRPO, a policy optimization method for grouped RL fine-tuning, where policies are updated based on grouped preferences over multiple rollouts.

\subsection{Consensus Prior Construction}

As illustrated in \autoref{fig:sample}, the construction of our consensus prior follows a two-stage pipeline: multi-source selection and consensus weighting.

\paragraph{Multi-source Selection} To identify candidate evidence frames, we combine signals from complementary sources, as depicted in \autoref{fig:sample}(A). To capture scene transitions, we compute the scene-cut prior shown in \autoref{fig:sample}(a1) by analyzing the Bhattacharyya distance between HSV histograms of adjacent sampled frames. Peaks in this metric signify substantial visual shifts, and a preset scene-change criterion is applied to isolate distinct scene boundaries from minor fluctuations. Additionally, we integrate the semantic prior in \autoref{fig:sample}(a2), which is derived from a pretrained visual-language encoder. By evaluating image-text similarity between sampled frames and the user query, we prioritize the top-ranked frames that are most directly relevant to the query. The selection process is grounded by the uniform baseline in \autoref{fig:sample}(a3), which maintains temporal coverage across the video.

\paragraph{Consensus Weighting \& Normalization} The selected signals are fused into a probabilistic distribution, as shown in \autoref{fig:sample}(B).
We assign importance weights $w_i$ to each frame $i$ based on the contributions from the uniform base, scene transitions, and semantic relevance, as shown in \autoref{fig:sample}(b1, b2):
\begin{equation}
w_i = w_{\text{base}} + \mathbb{I}(i \in \mathcal{S}_{\text{scene}}) \cdot \lambda_{\text{scene}} + \mathbb{I}(i \in \mathcal{S}_{\text{sem}}) \cdot \lambda_{\text{sem}},
\end{equation}
where $w_{\text{base}}$ prevents probability collapse, and $w_{\text{base}}$, $\lambda_{\text{scene}}$, and $\lambda_{\text{sem}}$ are hyperparameters.
These weights are then normalized to produce the final consensus prior $P(i)$, as shown in \autoref{fig:sample}(b3, b4):
\begin{equation}
P(i) = \frac{w_i}{\sum_{j=1}^{K} w_j}.
\end{equation}
This resulting distribution $P(i)$ serves as a temporal-annotation-free evidence prior for reward construction.

\begin{figure}[!t]
  \begin{center}
    \centerline{\includegraphics[width=\columnwidth]{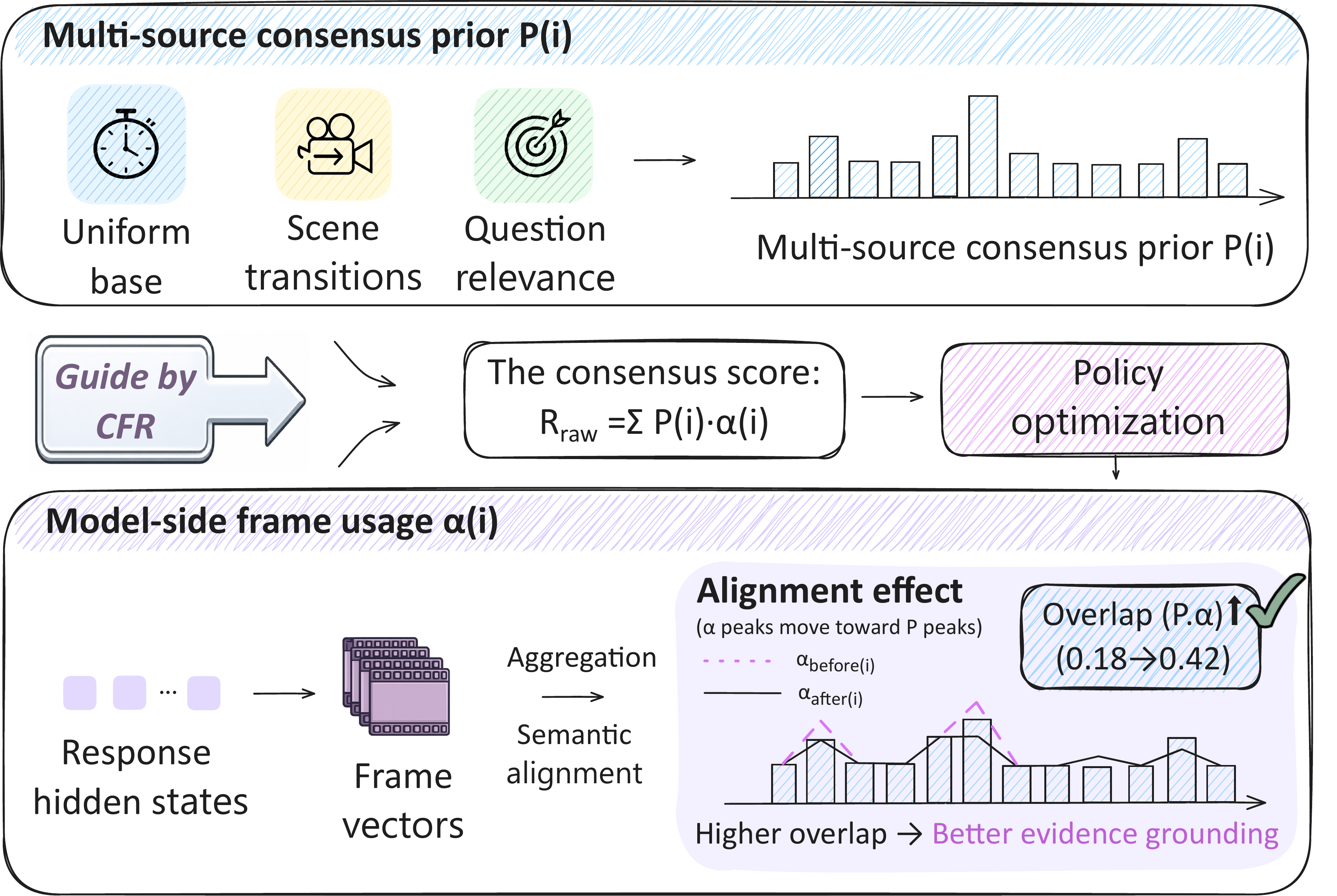}}
    \caption{Consensus Frame Reward as a process signal. CFR estimates a model-side frame-use distribution from response-frame similarity and rewards its overlap with the consensus prior $P(i)$. The resulting scalar reward guides GRPO toward generation behavior that is more aligned with candidate evidence frames.}
    \label{fig:cfr}
  \end{center}
\end{figure}

\subsection{Model-side Frame-use Score Extraction and Sharpening}
To align generation with the consensus prior, we require a frame-level score that is comparable to $P(i)$. We compute this model-side frame-use score from the similarity between visual frame representations and response hidden states. Since the model processes visual information as a sequence of patch tokens, these fine-grained signals are aggregated into frame-level vectors before the similarity calculation.
We aggregate frame features using salience-aware sparse aggregation to preserve sparse signals. Each sampled frame is encoded into $M$ visual patch tokens, yielding $H_{\text{vis}} \in \mathbb{R}^{K \times M \times D}$. The aggregated frame vector is computed by channel-wise max pooling:
\begin{equation}
[\mathbf{h}_i^v]_d = \max_{m=1}^{M} H_{\text{vis}}[i,m,d], \quad d=1,\ldots,D,
\end{equation}
capturing high-response regions without dilution by background noise.

For alignment, given the response hidden states $H_{\text{res}} \in \mathbb{R}^{L \times D}$, we compute raw weight scores:
\begin{equation}
S_{i} = \frac{1}{L} \sum_{t=1}^{L} \frac{\mathbf{h}_t^{\text{res}} \cdot (\mathbf{h}_i^v)^\top}{\|\mathbf{h}_t^{\text{res}}\| \|\mathbf{h}_i^v\|}.
\end{equation}
Here, $S_i$ quantifies the representation-level association between the $i$-th frame and the generated response. By averaging the cosine similarity across all response tokens $t$, we obtain a global frame-use score for the entire generation process.

The raw scores are sharpened with temperature $\tau$ by applying a softmax over the $K$ sampled frames:
\begin{equation}
\boldsymbol{w}_{\text{attn}}(i) =
\frac{\exp(S_i/\tau)}{\sum_{j=1}^{K}\exp(S_j/\tau)}.
\end{equation}
This enhances the signal-to-noise ratio for reliable policy gradients.

\paragraph{Consensus Frame Reward.} As shown in \autoref{fig:cfr}, the consensus frame reward is the dot product:
\begin{equation}
R_{\text{cf}} = \mathbf{P} \cdot \boldsymbol{w}_{\text{attn}} = \sum_{i=1}^{K} P(i) \cdot \boldsymbol{w}_{\text{attn}}(i),
\end{equation}
rewarding agreement between the temporal-annotation-free consensus prior and the model-side frame-use score. Because both vectors are normalized over the $K$ sampled frames, the raw overlap is a bounded scalar whose magnitude depends on the frame budget; when visualizing training dynamics, we therefore report a scaled version of this overlap as a diagnostic rather than as a separate reward term.

\subsection{Reward Design and Optimization}
The final component of our framework is the optimization objective, shown in \autoref{fig:main} (B). We use a composite reward function that preserves standard outcome supervision while adding evidence-level feedback. The reward contains four terms: accuracy for answer correctness, structural constraints for response format and conciseness, temporal consistency for ordered video reasoning, and CFR for agreement with candidate evidence frames.
The total reward integrates these standard and process-oriented terms:
\begin{equation}
R_{\text{total}}(o) = R_{\text{acc}} + R_{\text{struct}} + \lambda \cdot R_{\text{cf}} + R_{\text{temp}},
\end{equation}
where $R_{\text{acc}}$ evaluates answer accuracy, and $R_{\text{struct}}$ (format-and-length reward) penalizes excessive generation length and incorrect response formats to encourage conciseness and structural compliance. The temporal reward $R_{\text{temp}}$ encourages sensitivity to frame order:
\begin{equation}
R_{\text{temp}} =
\begin{cases}
\gamma & \text{if } \text{Acc}_{\text{ordered}} \geq \text{Acc}_{\text{shuffled}} \text{ and } o \text{ is correct} \\
0 & \text{otherwise},
\end{cases}
\end{equation}
Here, $\text{Acc}_{\text{ordered}}$ and $\text{Acc}_{\text{shuffled}}$ denote verifiable answer scores obtained from the ordered and shuffled frame sequences for the same sample. This reward incentivizes the model to perform better on temporally ordered inputs than on shuffled sequences when the answer is correct.

Optimization uses GRPO \cite{shao2024deepseekmath}, grouping rollouts and updating the policy $\pi$ via preference optimization. Let
\begin{equation}
\rho_{i,t}(\theta)=
\frac{\pi_{\theta}(o_{i,t} \mid v,q,o_{i,<t})}
{\pi_{\theta_{\text{old}}}(o_{i,t} \mid v,q,o_{i,<t})},
\end{equation}
where $v$ denotes the sampled video input and $\mathcal{D}$ denotes the training distribution. For each video-question pair $(v,q)$, a group of outputs is sampled from the old policy. Following GRPO, the clipped objective is:
\begin{equation}
\begin{aligned}
    \mathcal{J}_{\text{CF-GRPO}}(\theta) &=
    \mathbb{E}_{\mathcal{D},\,\pi_{\theta_{\text{old}}}}
    \left[
    \frac{1}{G}\sum_{i=1}^{G}
    \frac{1}{|o_i|}\sum_{t=1}^{|o_i|}
    \ell_{i,t}(\theta)
    \right], \\
    \ell_{i,t}(\theta) &=
    \min \big(
    \rho_{i,t}(\theta)\hat{A}_{i},
    \operatorname{clip}_{\epsilon}(\rho_{i,t}(\theta))\hat{A}_{i}
    \big) \\
    &\quad
    - \beta\,\mathbb{D}_{\mathrm{KL}}
    [\pi_{\theta}\| \pi_{\mathrm{ref}}]_{i,t},
\end{aligned}
\end{equation}
where $\operatorname{clip}_{\epsilon}(x)=\operatorname{clip}(x,1-\epsilon,1+\epsilon)$, and $\hat{A}_{i}$ is estimated from the group-normalized completion reward based on $r(o_i) = R_{\text{total}}(o_i)$. The term $\mathbb{D}_{\mathrm{KL}}[\pi_{\theta}\| \pi_{\mathrm{ref}}]_{i,t}$ denotes the token-level KL estimator used in GRPO. Thus, the policy update is token-wise, while the reward and advantage are defined for the complete rollout. This loop optimizes a consensus-aware policy while keeping the model architecture unchanged.

\section{Experiments}

\subsection{Experimental Setup}
\paragraph{Training Details}
Training begins with one epoch of SFT on the Video-R1 CoT-165k dataset and continues with VideoCFR optimization using the proposed CF-GRPO algorithm on 8 NVIDIA A800 GPUs with a per-device batch size of 4. The training data incorporates both image and video reasoning samples. The video portion comprises 100K video-question-answer pairs sampled from MSRVTT-QA, MSVD-QA, and ActivityNet-QA, augmented with synthetic temporal reasoning queries. For efficiency, training limits video frames to a maximum of 16, with each frame processed under a dynamically adapted pixel budget up to $128 \times 28 \times 28$. During inference, the pixel budget increases to $256 \times 28 \times 28$, and the sampled frame budget ranges from 16 to 64 for better performance. Group size $G$ is set to 8. Training runs for 5 epochs, taking approximately 48 hours.

\paragraph{Implementation Hyperparameters}
We use the AdamW optimizer with a learning rate of $5 \times 10^{-6}$ and weight decay of 0.01. The KL divergence hyperparameter $\beta$ is 0.04. Gradient norm is clipped to 5 for stability. For the CFR mechanism, the reward scaling factor is $\lambda=3.0$, and the sharpening temperature is $\tau=0.1$. Hierarchical prior weights are $w_{\text{base}}=0.1$, $\lambda_{\text{scene}}=1.0$, and $\lambda_{\text{sem}}=3.0$. The temporal reward parameter is $\gamma = 0.3$.

\paragraph{Benchmarks and Baselines}
We evaluate on a suite of video reasoning benchmarks (VSI-Bench \cite{yang2025thinking}, VideoMMMU \cite{hu2025video}, MMVU(MC) \cite{zhao2025mmvu}) and general video understanding benchmarks (MVBench \cite{li2024mvbench}, TempCompass \cite{liu2024tempcompass}, VideoMME(w/o sub) \cite{fu2025video}). The comparison includes standard Video-MLLMs and recent RL-enhanced baselines when their reported metrics match these benchmarks.

\subsection{Main Results}
We compare VideoCFR with representative Video-MLLMs and recent RL-enhanced baselines in \autoref{tab:main_results}. The table separates standard Video-MLLMs from RL-based methods to make the comparison axes clear. Overall, VideoCFR obtains competitive results across both reasoning-oriented and general video-understanding benchmarks, suggesting that adding an evidence-level CFR signal is useful beyond answer-only supervision.

\begin{table*}[!t]
  \caption{Video model performance comparison. Results are reported as accuracy (\%). For MARC-3B, the frame entry follows the original paper and denotes one-frame-equivalent visual tokens after compression.}
  \label{tab:main_results}
  \begin{center}
    \begin{small}
      \begin{sc}
      \resizebox{\textwidth}{!}{
            \begin{tabular}{lcccc|ccc}
              \toprule
              && \multicolumn{3}{c}{Video Reasoning Benchmark} & \multicolumn{3}{c}{Video General Benchmark} \\
              \cmidrule(r){3-5} \cmidrule(l){6-8}
              Models & Frames & VSI-Bench & VideoMMMU & MMVU(MC) & MVBench & TempCompass & VideoMME(w/o sub) \\
              \midrule
              LLaMA-VID \cite{Llama-vid} & - & - & - & - & 41.9 & 45.6 & - \\
              VideoLLaMA2 \cite{VIDEOLLAMA2} & - & - & - & 44.8 & 54.6 & - & 47.9 \\
              LongVA-7B \cite{LONGVA} & - & 29.2 & 23.9 & - & - & 56.9 & 52.6 \\
              VILA-1.5-8B \cite{VILA-1.5} & - & 28.9 & 20.8 & - & - & 58.8 & - \\
              VILA-1.5-40B \cite{VILA-1.5} & - & 31.2 & 34.0 & - & - & - & 60.1 \\
              Video-UTR-7B \cite{VIDEO-UTR} & - & - & - & - & 58.8 & 59.7 & 52.6 \\
              LLaVA-OneVision-7B \cite{LLAVA-ONEVISION} & - & 32.4 & 33.8 & 49.2 & 56.7 & - & 58.2 \\
              Kangeroo-8B \cite{KANGEROO} & - & - & - & - & 61.1 & 62.5 & 56.0 \\
              Qwen2.5-VL-7B \cite{Qwen2.5-VL} & - & - & 47.4 & 61.3 & 59.4 & 69.2 & 52.8 \\
              \midrule
              Video-R1-7B \cite{videor1} & 16 & 30.3 & 47.2 & 63.5 & 62.4 & 70.8 & 54.3 \\
              VideoChat-R1 \cite{li2025videochat} & 16 & 28.9 & 48.7 & 65.8 & 64.2 & \textbf{73.5} & 57.7 \\
              DeepVideo-R1 \cite{park2025deepvideo} & - & 33.0 & 40.7 & 59.0 & 49.6 & 63.1 & 51.1 \\
              TinyLLaVA-Video-R1 \cite{TinyLLaVA-Video-R1} & 16 & - & - & 46.9 & - & 49.5 & 46.6 \\
              VIDEORFT \cite{VIDEORFT} & 32 & - & - & 51.1 & 62.1 & - & - \\
              Temporal-RLT \cite{Temporal-RLT} & 32 & - & - & 65.0 & - & - & 57.6 \\
              MARC-3B \cite{wu2026marc} & 1 & 27.6 & 33.1 & 52.0 & 45.8 & 55.3 & 39.4 \\
              \midrule
              \rowcolor{oursrow} VideoCFR (Ours) & 16 & 31.8 & 50.5 & 66.4 & \textbf{66.1} & 70.8 & 55.1 \\
              \rowcolor{oursrow} VideoCFR (Ours) & 32 & 33.1 & \textbf{52.4} & 65.9 & 64.5 & 72.8 & 58.9 \\
              \rowcolor{oursrow} VideoCFR (Ours) & 64 & \textbf{34.8} & 50.6 & \textbf{66.7} & 63.9 & 72.9 & \textbf{61.1} \\
              \bottomrule
            \end{tabular}
        }
      \end{sc}
    \end{small}
  \end{center}
\end{table*}

\paragraph{Comparison with Video-MLLM and RL Baselines}
Existing RL approaches, such as Video-R1-7B, VideoChat-R1, and Temporal-RLT, already improve over many standard baselines by introducing video-specific reward designs. VideoCFR further improves several benchmark scores under comparable model scale, suggesting that frame-level evidence alignment provides an additional training signal. For example, on VideoMMMU, our 32-frame model achieves 52.4\%, improving over the reported Video-R1-7B result by 5.2\% and the reported VideoChat-R1 result by 3.7\%. Compared with the recent MARC-3B compression model, VideoCFR reports higher accuracy on all six benchmarks, although this comparison should be interpreted as accuracy context because MARC-3B is optimized for one-frame-equivalent token compression with a smaller backbone. These results support the role of evidence-level reward design in Video-MLLM post-training, while the per-benchmark gains remain dependent on frame budget and task type.

\paragraph{Scaling with Temporal Context}
Increasing the sampled frame budget changes performance in a benchmark-dependent manner. VideoCFR benefits from more frames on several benchmarks: increasing from 16 to 64 frames improves VideoMME (55.1\% $\rightarrow$ 61.1\%), VSI-Bench (31.8\% $\rightarrow$ 34.8\%), and TempCompass (70.8\% $\rightarrow$ 72.9\%). However, VideoMMMU and MVBench peak at smaller frame budgets. This pattern suggests that CFR can help the model use additional temporal context when the benchmark benefits from richer evidence, but additional frames are not uniformly beneficial across all tasks.

\subsection{Ablation and Control Studies}
We separate reward-component ablations from broader training and data controls. \autoref{tab:ablation} isolates the components inside the CFR mechanism under the same 16-frame VideoCFR setting, while \autoref{tab:extended_ablation} varies the training recipe, the use of image QA data, and the use of an external frame-selection method. This separation avoids conflating internal reward design with changes in training stages or input-frame selection.

\begin{table*}[!t]
  \caption{Ablation study of VideoCFR reward components. Results are reported as accuracy (\%).}
  \label{tab:ablation}
  \begin{center}
    \begin{small}
      \begin{sc}
        \resizebox{\textwidth}{!}{
            \begin{tabular}{lcccc|ccc}
              \toprule
              && \multicolumn{3}{c}{Video Reasoning Benchmark} & \multicolumn{3}{c}{Video General Benchmark} \\
              \cmidrule(r){3-5} \cmidrule(l){6-8}
              Models & Frames & VSI-Bench & VideoMMMU & MMVU(MC) & MVBench & TempCompass & VideoMME(w/o sub) \\
              \midrule
              \rowcolor{oursrow} VideoCFR (Ours) & 16 & 31.8 & 50.5 & 66.4 & 66.1 & 70.8 & 55.1 \\
              Uniform Prior & 16 & 31.4 & 47.3 & 63.5 & 63.1 & 71.0 & 55.8 \\
              w/o Semantic Prior & 16 & 31.8 & 48.9 & 64.3 & 64.1 & 70.9 & 54.4 \\
              w/o Scene Prior & 16 & 31.0 & 46.1 & 65.8 & 63.2 & 70.7 & 54.1 \\
              w/o Sparse Aggregation & 16 & 32.2 & 47.3 & 64.9 & 63.6 & 70.8 & 55.4 \\
              w/o Sharpening & 16 & 30.6 & 49.1 & 64.9 & 63.2 & 71.1 & 54.2 \\
              \bottomrule
            \end{tabular}
         }
      \end{sc}
    \end{small}
  \end{center}
\end{table*}

\paragraph{Reward-Component Ablation}
The ablation starts from the source of the reward signal by contrasting the multi-source consensus prior with a uniform prior. The uniform prior assumes an equiprobable distribution across frames ($P(i) = 1/K$). Since this makes the overlap with any normalized model-side frame-use distribution constant, this variant removes the informative frame-prior signal while preserving the same training setting. In contrast, our consensus prior integrates scene-transition and semantic relevance cues into a non-uniform evidence prior. As detailed in \autoref{tab:ablation}, replacing the consensus prior with a uniform distribution reduces all three reasoning benchmarks and MVBench, most notably by 3.2\% on VideoMMMU, although it yields small gains on TempCompass and VideoMME. Removing the semantic prior reduces four of six benchmarks, while removing the scene prior reduces all six benchmarks. These results indicate that the consensus prior is most beneficial on reasoning-oriented evaluations, and that combining complementary evidence sources generally provides a stronger signal than relying on a single cue.

The same table also evaluates the model-side signal construction. Removing sparse aggregation lowers VideoMMMU, MMVU(MC), and MVBench, which are benchmarks where localized or fine-grained visual evidence can affect the answer. Removing distribution sharpening reduces most reasoning and general benchmarks, including VSI-Bench, VideoMMMU, MMVU(MC), MVBench, and VideoMME. The remaining small metric gains in a few columns indicate that these components are not uniformly beneficial for every benchmark, but the overall pattern supports their role in making CFR more informative for complex reasoning tasks.

\begin{table*}[!t]
  \caption{Control study on training stages, training data, and external frame selection. Results are reported as accuracy (\%).}
  \label{tab:extended_ablation}
  \begin{center}
    \begin{small}
      \begin{sc}
        \resizebox{\textwidth}{!}{
            \begin{tabular}{lcccc|ccc}
              \toprule
              && \multicolumn{3}{c}{Video Reasoning Benchmark} & \multicolumn{3}{c}{Video General Benchmark} \\
              \cmidrule(r){3-5} \cmidrule(l){6-8}
              Variant & Frames & VSI-Bench & VideoMMMU & MMVU(MC) & MVBench & TempCompass & VideoMME(w/o sub) \\
              \midrule
              SFT & 16 & 30.2 & 44.6 & 59.2 & 57.1 & 69.4 & 51.9 \\
              SFT + GRPO & 16 & 32.7 & 48.3 & 62.1 & 61.1 & 71.3 & 54.5 \\
              SFT + GRPO + Temporal & 16 & 31.2 & 46.8 & 63.2 & 60.8 & 71.1 & 53.4 \\
              VideoCFR w/o SFT & 16 & 31.8 & 49.5 & 63.8 & 60.4 & 70.9 & 53.8 \\
              SFT + AKS & 16 & 32.6 & 45.0 & 59.0 & 57.8 & 65.9 & 51.9 \\
              \rowcolor{oursrow} VideoCFR & 16 & 31.8 & 50.5 & 66.4 & 66.1 & 70.8 & 55.1 \\
              VideoCFR + AKS & 16 & 31.7 & 46.9 & 63.5 & 63.4 & 69.4 & 55.3 \\
              VideoCFR w/o Image-QA & 16 & 31.9 & 43.4 & 61.4 & 63.9 & 70.5 & 54.1 \\
              \bottomrule
            \end{tabular}
         }
      \end{sc}
    \end{small}
  \end{center}
\end{table*}

\paragraph{Training, Data, and Frame-Selection Controls}
We further evaluate the effects of training stages, training data, and external frame selection in \autoref{tab:extended_ablation}. SFT only applies cold-start supervised fine-tuning to the base model. SFT + GRPO is a pure reinforcement-learning control without temporal or consensus-frame rewards, while SFT + GRPO + Temporal retains the temporal reward but excludes the consensus-frame reward. VideoCFR w/o SFT applies the VideoCFR reward design directly to the base model without cold-start supervised fine-tuning. VideoCFR improves over SFT on all six benchmarks, over SFT + GRPO + Temporal on five of six benchmarks, matches or improves over VideoCFR w/o SFT on five of six benchmarks, and improves over SFT + GRPO on four of six benchmarks. These results indicate that cold-start SFT and CFR-based RL provide complementary gains while the effect remains benchmark-dependent.

AKS \cite{tang2025adaptive} is used as an external frame-selection method whose selected frames are fed into the same model. Under the matched 16-frame setting, AKS changes performance in a benchmark-dependent manner: it improves some SFT results but does not reproduce the gains obtained by VideoCFR, especially on VideoMMMU, MMVU(MC), and MVBench. This suggests that the gains of VideoCFR cannot be explained by input-side frame selection alone. VideoCFR w/o Image-QA trains without image QA data and uses only video QA, leading to clear drops on VideoMMMU and MMVU(MC), which indicates that image QA data remains useful for preserving fine-grained visual recognition during video RL training.

\subsection{Diagnostic Analysis}
The following analyses examine whether the internal signals of CFR behave consistently with the intended evidence-alignment mechanism. They do not introduce additional training variants; instead, they inspect training dynamics, frame-use distributions, spatial focus, and attention-event concentration.

\begin{figure}[!t]
  \begin{center}
    \centerline{\includegraphics[width=\columnwidth]{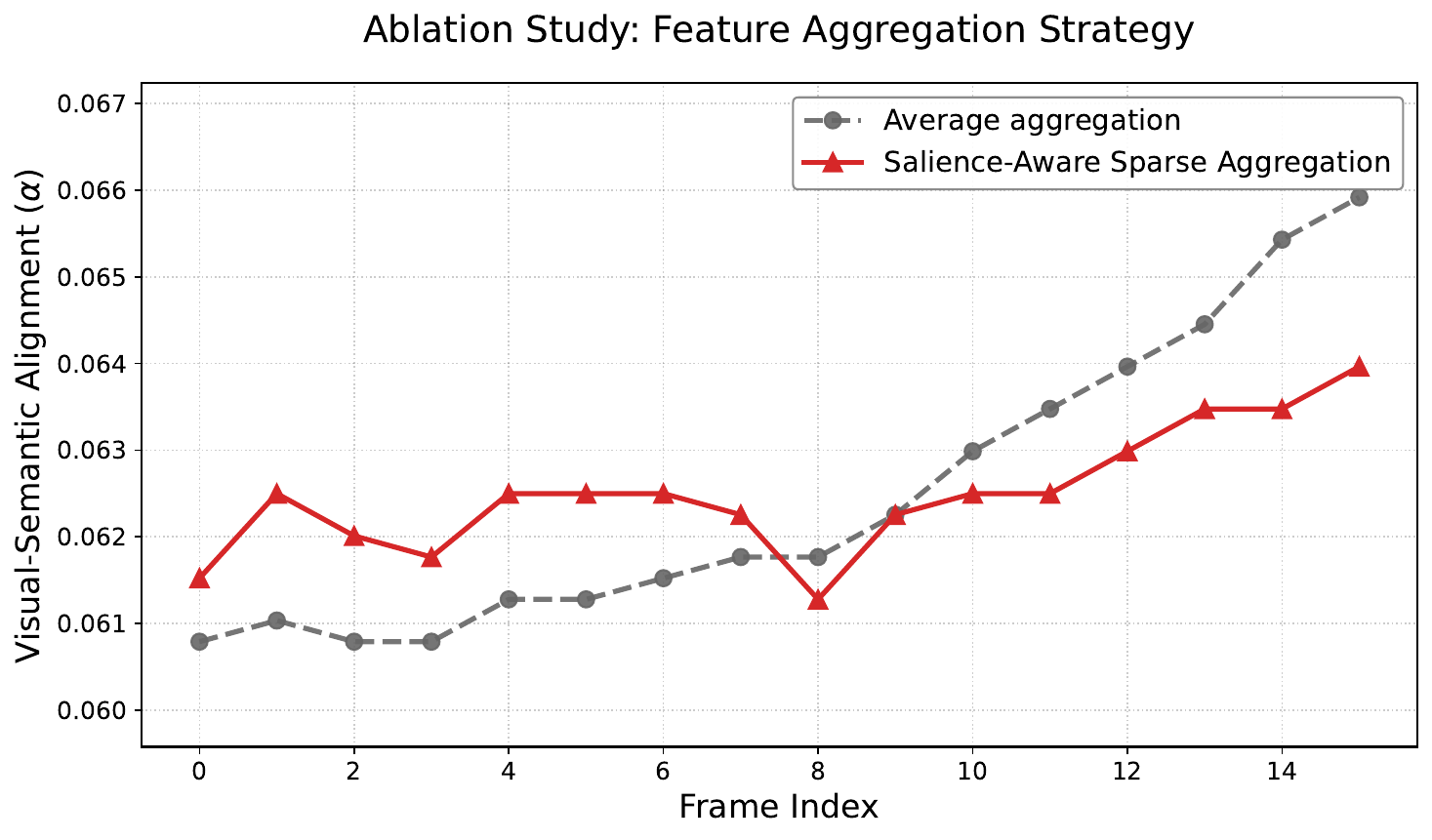}}
    \caption{Effect of feature aggregation. Salience-aware sparse aggregation better preserves sparse high-response cues than average aggregation, yielding more discriminative frame-level alignment.}
    \label{fig:ablation_pooling}
  \end{center}
\end{figure}

\paragraph{Effect of Sparse Feature Aggregation}
We analyze the feature aggregation strategy used to extract frame-level semantics from sparse visual tokens. Average aggregation can attenuate high-activation visual features by mixing them with background tokens, while salience-aware sparse aggregation preserves the strongest local responses for reward computation. As illustrated in \autoref{fig:ablation_pooling}, average aggregation shows a systematic temporal bias and misses early-stage visual semantics. In contrast, salience-aware sparse aggregation maintains sharper responses throughout the sequence and exhibits distinct peaks at information-rich moments such as Frames 1 and 4--6. This supports its role in amplifying task-relevant signals before computing the model-side frame-use score.

\begin{figure}[!t]
  \begin{center}
    \centerline{\includegraphics[width=\columnwidth]{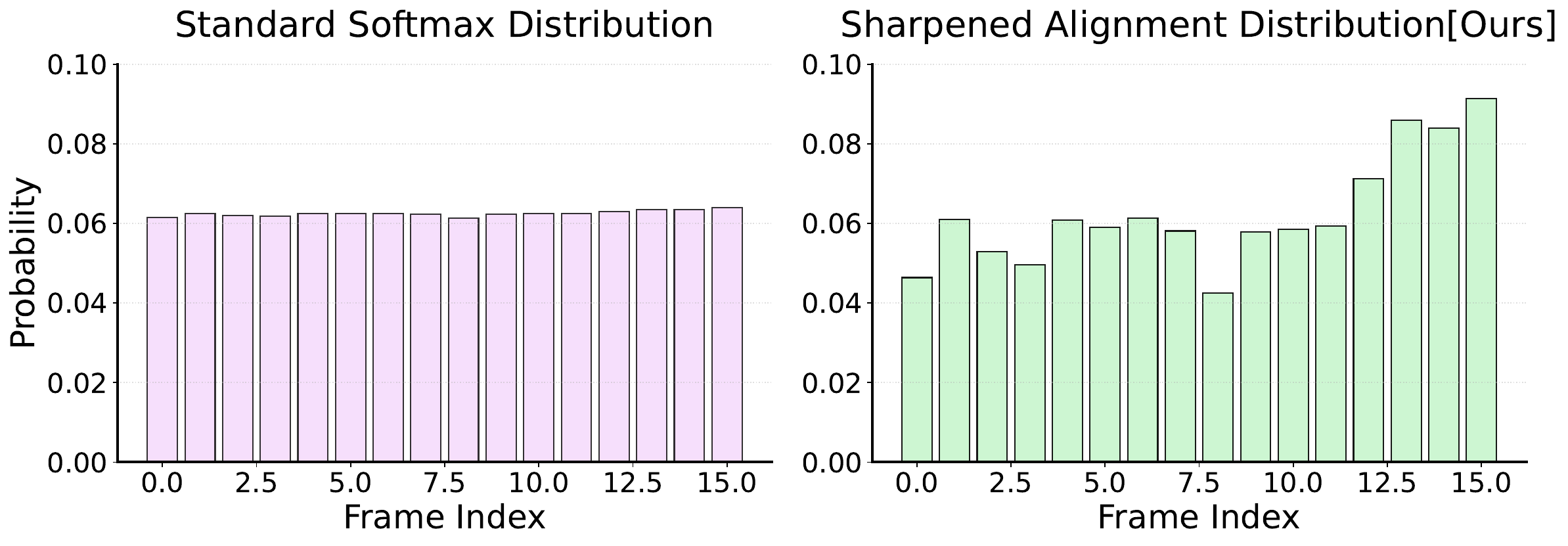}}
    \caption{Effect of distribution sharpening. Low-temperature sharpening produces a more peaked frame-use distribution than a high-entropy softmax, strengthening the process signal for optimization.}
    \label{fig:ablation_temperature}
  \end{center}
\end{figure}

\begin{figure}[!t]
  \centering
  \includegraphics[width=\columnwidth]{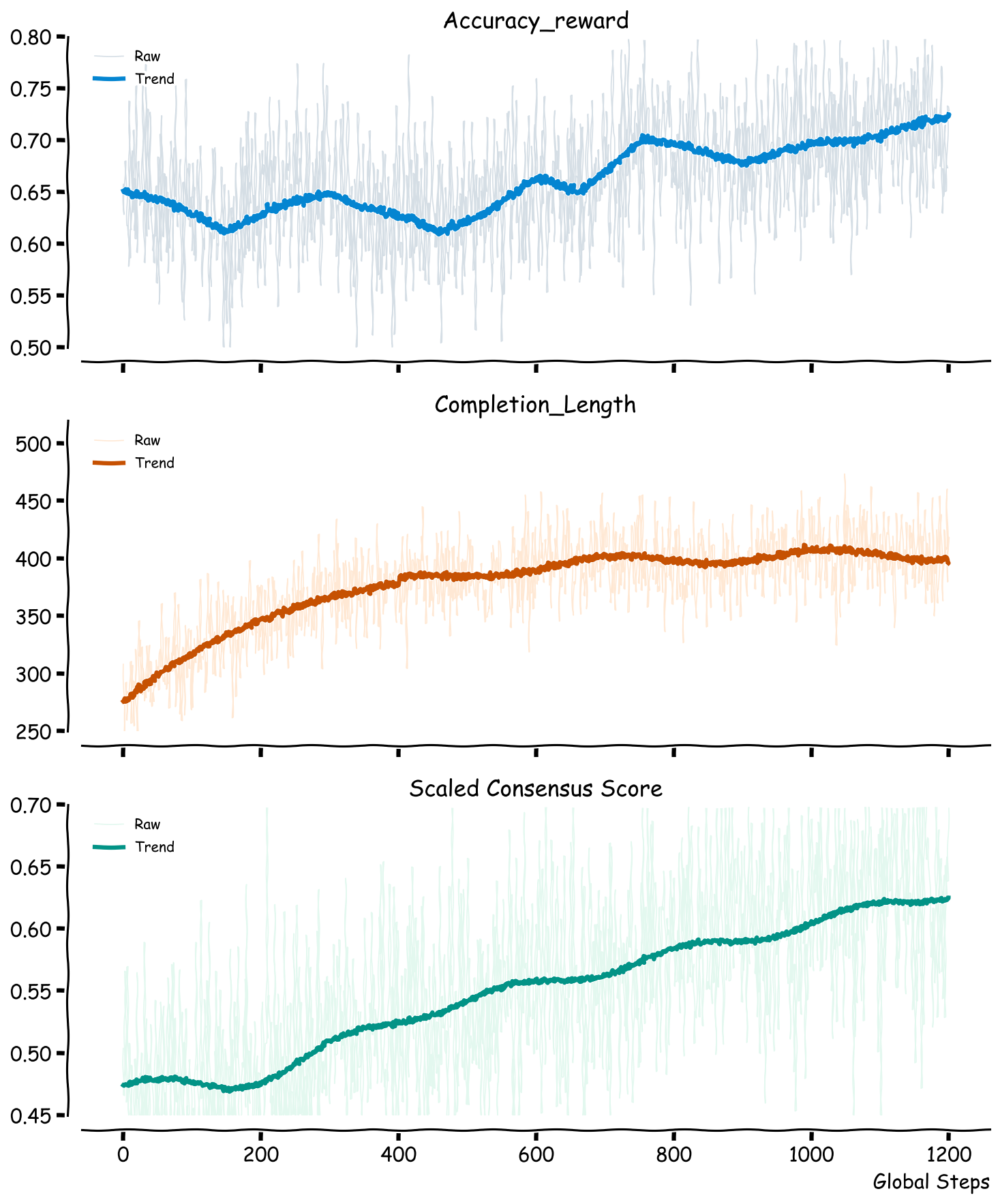}
  \caption{Training dynamics of VideoCFR-7B. The curves show the progression of reasoning accuracy, completion length, and scaled consensus alignment during RL training.}
  \label{fig:reward_curves}
\end{figure}
\begin{figure}[!t]
  \begin{center}
    \centerline{\includegraphics[width=\columnwidth]{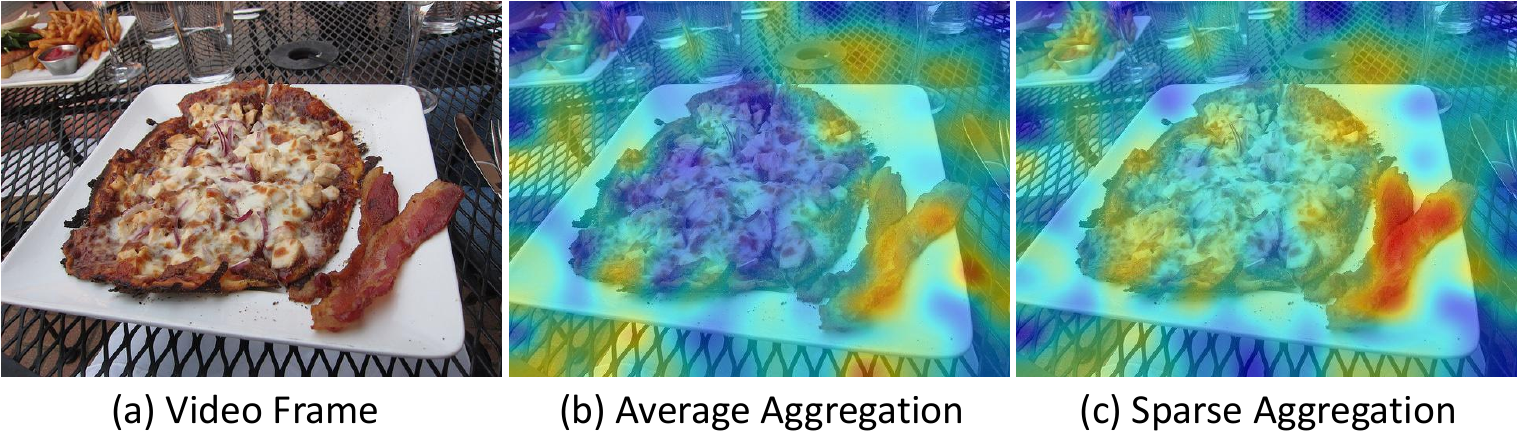}}
    \caption{Spatial focus visualization on multiple objects. (a) Video frame. (b) Average aggregation heatmap showing fragmented attention. (c) Salience-aware sparse aggregation heatmap covering the pizza and bacon while retaining attention on the background fries.}
    \label{fig:pooling-att}
  \end{center}
\end{figure}

\begin{figure*}[!t]
  \centering
  \includegraphics[width=0.8\textwidth]{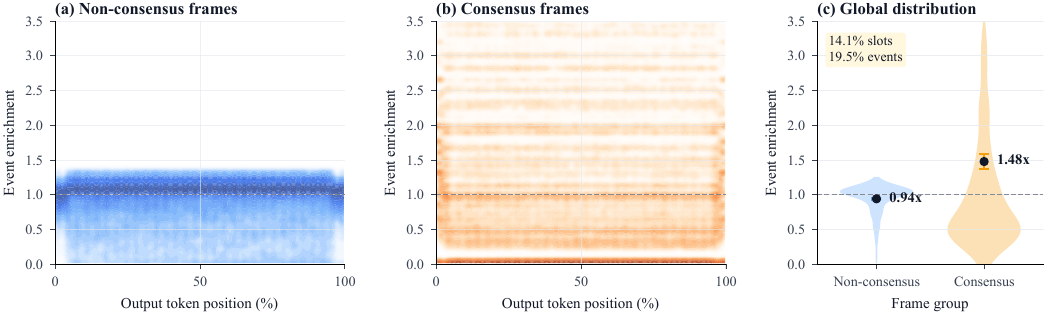}
  \caption{Top-attention event enrichment over normalized output-token positions. (a) Density distribution for non-consensus frames. (b) Density distribution for consensus frames. (c) Global enrichment statistics across frame groups. Enrichment is measured relative to the frame-slot baseline, where $1.0\times$ denotes proportional allocation.}
  \label{fig:attention_density}
\end{figure*}

\paragraph{Effect of Distribution Sharpening}
We examine distribution sharpening by comparing the low-temperature softmax with a standard softmax. As visualized in \autoref{fig:ablation_temperature}, the standard softmax produces a more diffuse distribution, which weakens the contrast among candidate frames. The low-temperature variant provides a higher-contrast frame-use distribution. This is important for CFR because a nearly uniform model-side distribution would make the overlap with the consensus prior less discriminative and would provide weaker policy-gradient feedback.

\paragraph{Training Dynamics and Consensus Alignment}
We visualize the training dynamics of VideoCFR-7B in \autoref{fig:reward_curves}. The accuracy reward increases and stabilizes during training, while completion length gradually grows and then plateaus. The scaled consensus score rises from 0.47 to over 0.62, indicating that the CFR term is not merely a static auxiliary score but is optimized during RL training. The simultaneous stabilization of accuracy and completion length suggests that the increase in consensus alignment is not simply caused by unbounded response-length growth. Completion length is still reported only as a training diagnostic and should not be interpreted as a direct measure of reasoning quality.

\paragraph{Spatial Focus under Sparse Aggregation}
To inspect the learned frame-use behavior at the spatial level, we visualize heatmaps for a food-recognition query in \autoref{fig:pooling-att}. The visualization shows higher response-associated scores on the primary objects---bacon and pizza---while retaining sensitivity to peripheral details such as the fries in the background. Compared with average aggregation, salience-aware sparse aggregation produces a less fragmented pattern and suppresses several background regions. This pattern supports the intended role of sparse aggregation: it preserves localized high-response visual regions that may be diluted by averaging, so the frame-use score used by CFR is less dominated by background tokens.

\paragraph{Attention-Event Enrichment on Consensus Frames.}
We further analyze whether high-attention response-frame events are concentrated on consensus frames. For each output-token position, we collect global top-5\% frame-token attention events and compute their enrichment relative to the corresponding frame-slot baseline, where $1.0\times$ indicates proportional allocation. This metric asks whether consensus frames receive more high-attention events than expected from their share of available frame slots.

As shown in \autoref{fig:attention_density}(a), non-consensus frames mostly concentrate around or below this baseline over normalized output-token positions. In contrast, \autoref{fig:attention_density}(b) shows that consensus frames cover a broader above-baseline region, indicating that high-attention events are more likely to occur on frames favored by the consensus prior. The aggregate statistics in \autoref{fig:attention_density}(c) provide the same pattern at the frame-group level: consensus frames occupy 14.1\% of frame slots but account for 19.5\% of top-attention events, with a mean enrichment of $1.48\times$, whereas non-consensus frames remain below baseline with a mean enrichment of $0.94\times$. These results do not replace quantitative benchmark evaluation, but they are consistent with the interpretation that CFR encourages response generation to align with frames identified as candidate visual evidence.

\subsection{Qualitative Case Studies}
\label{sec:qualitative_case_studies}

To connect the diagnostic patterns above with concrete answer behavior, we present qualitative cases. These cases are not used as substitutes for benchmark evaluation; they illustrate how evidence alignment can affect model focus when the relevant visual cue is localized, temporally brief, or visually confusable with irrelevant content.

\begin{figure*}[!t]
  \centering
  \includegraphics[width=0.8\textwidth]{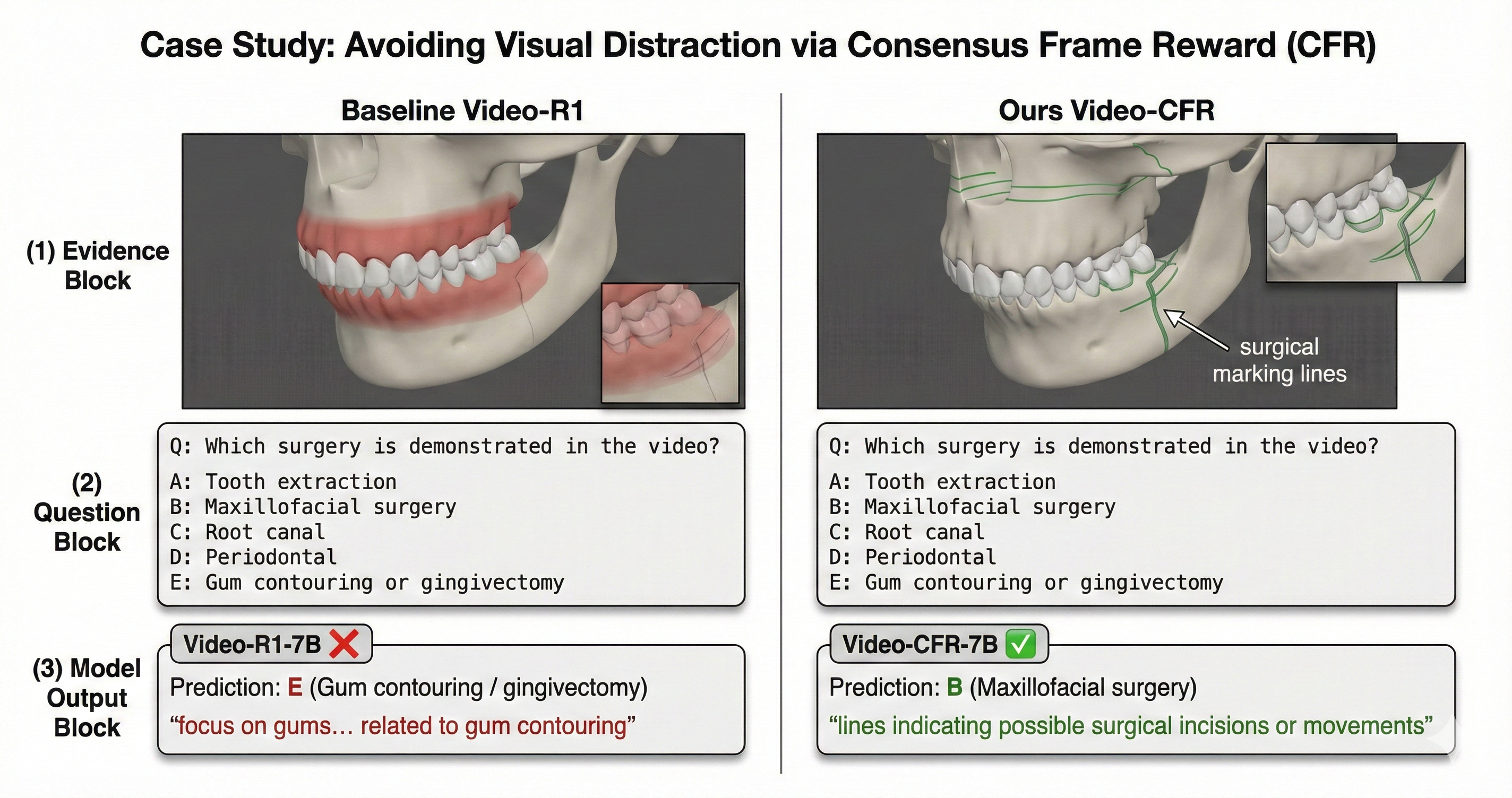}
  \caption{Case study on visual distraction. The baseline (Video-R1) attends to the salient but irrelevant gums and predicts an incorrect option. VideoCFR assigns higher evidence to the surgical marking lines and predicts the correct procedure.}
  \label{fig:case_study1}
\end{figure*}

\paragraph{Robustness to Visual Distractions}
\autoref{fig:case_study1} examines a medical visualization in which the relevant cue is spatially localized. The baseline model (Video-R1-7B) focuses on the gums, which are visually salient but not decisive for the queried procedure, and selects an incorrect procedure option. This behavior is consistent with a wrong-evidence failure mode: the response can be anchored by locally salient content and common language associations, such as linking teeth or gums with contouring, rather than by procedure-specific evidence.

\begin{figure*}[!t]
  \centering       
  \includegraphics[width=0.8\textwidth]{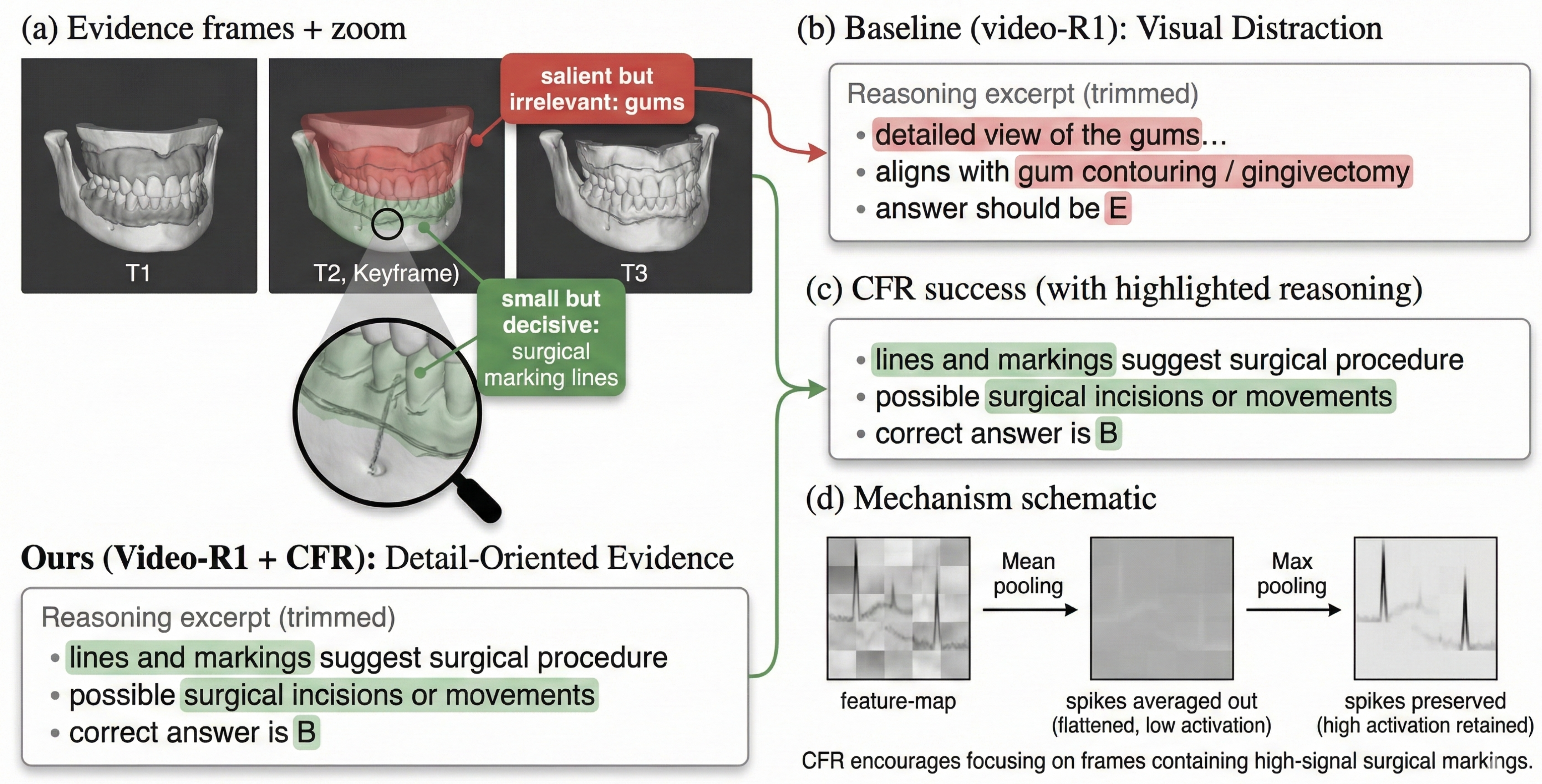}
  \caption{Mechanism schematic for evidence aggregation. The comparison of evidence frames and reasoning excerpts illustrates how salience-aware sparse aggregation preserves localized surgical markings, whereas mean pooling attenuates them through averaging.}
  \label{fig:case_study2}
\end{figure*}

\begin{figure*}[!t]
  \centering
  \includegraphics[width=\textwidth]{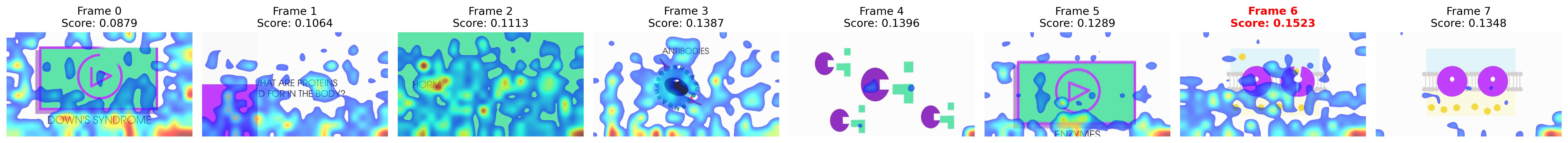}
  \caption{\textbf{Visualization of temporal attention and spatial saliency.} The top row displays representative consensus frames from the video with their corresponding attention scores. Higher scores correspond to frames containing task-relevant content (e.g., Frame 6 showing protein transport mechanisms), while lower scores are assigned to less relevant introductory frames. The heatmaps illustrate the model's spatial focus under salience-aware sparse aggregation.}
  \label{fig:sequence_attention_case}
\end{figure*}
\begin{figure*}[!t]
  \centering
  \includegraphics[width=0.8\textwidth]{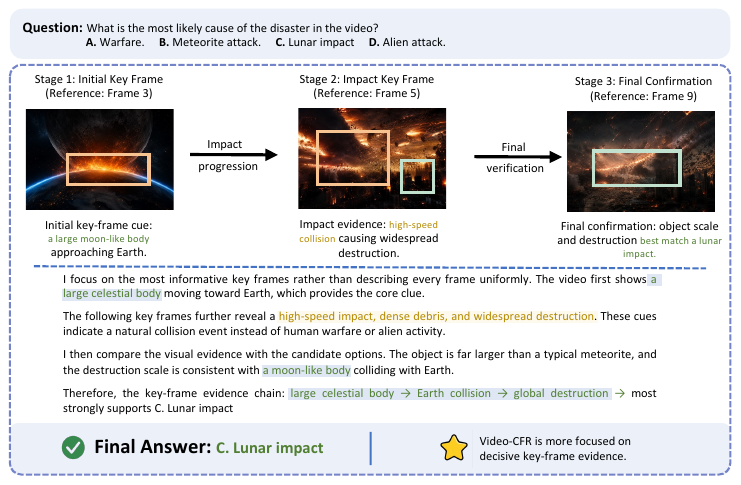}
  \caption{Qualitative case on key-frame evidence chaining. The visualized rationale links an initial moon-like object, the subsequent impact, and the final destruction pattern before selecting the lunar-impact answer.}
  \label{fig:case_cfr}
\end{figure*}

In contrast, VideoCFR-7B assigns higher evidence to the surgical marking lines and selects the correct procedure option. \autoref{fig:case_study2} provides a mechanism-level view of this behavior. Mean pooling can attenuate localized signals such as surgical markings by averaging them with background pixels, whereas sparse aggregation preserves localized high-response regions for reward computation. The consensus reward then encourages the model-side frame-use score to align with candidate evidence frames identified through scene-cut and semantic cues. This case is therefore consistent with the intended interaction between sparse signal preservation and consensus-frame alignment.

\paragraph{Temporal-Spatial Focusing Across a Sequence}
\autoref{fig:sequence_attention_case} examines whether the model uses non-uniform temporal evidence in a video sequence explaining biological mechanisms. The model assigns lower weight to less informative introductory content and higher weight to frames containing relevant mechanisms. Frame 0, an introductory title page, receives the lowest score (0.0879), while Frame 6, which depicts protein transport, receives the highest score (0.1523). This non-uniform distribution is consistent with the effect of distribution sharpening, which is intended to make candidate evidence frames more distinguishable.

The spatial heatmaps provide complementary evidence for localized visual focus. In Frame 3, high-activation regions concentrate on antibody structures and relevant textual annotations, while the white background receives lower activation. Together with the temporal peaks at Frames 3 and 6, this case suggests that VideoCFR can combine temporal frame selection with spatially localized evidence use when multiple frames support the target biological category. This behavior is aligned with the diagnostic results in \autoref{fig:attention_density}, where consensus frames receive above-baseline high-attention events.

\paragraph{Key-Frame Evidence Chaining}
\autoref{fig:case_cfr} examines a case where the answer depends on a small set of temporally separated visual cues. The selected evidence frames form a sequence that progresses from an approaching moon-like body through a high-speed collision to large-scale destruction. The corresponding rationale compares this evidence with the candidate options and selects the lunar-impact answer rather than distractor categories such as meteorite attack or warfare. This case is consistent with the qualitative role of CFR as an evidence-alignment signal: the answer is supported by a non-uniform chain of relevant frames instead of a frame-independent description of the video.

\section{Limitations}

Although CF-GRPO provides a process-level reward for evidence-aware video reasoning without human temporal annotations, it still has several limitations.

At the evidence-prior level, the consensus prior is an estimate of candidate evidence rather than ground-truth temporal supervision. It is constructed from temporal coverage, scene-transition cues, and query-conditioned visual relevance, so it may be less reliable when the necessary evidence is implicit, visually subtle, or distributed across many frames.
This constraint also limits the modality coverage of the current prior, which uses visual and text-query cues only. Videos that require audio, subtitles, speech content, or cross-modal event timing may require additional consensus signals beyond the frame-level visual evidence considered here.
At the optimization level, CF-GRPO introduces additional computation during training because it computes consensus priors, model-side frame-use scores, and RL updates over grouped rollouts. In addition, increasing the sampled frame budget is not uniformly beneficial across benchmarks, which suggests that frame-budget selection should be adapted to the task rather than treated as a universal scaling rule.
At the evaluation level, our diagnostic analyses use internal frame-use and attention-event statistics to inspect evidence alignment. These analyses help interpret the training behavior, but they are not a substitute for evaluations with human-annotated temporal evidence labels.

Future work could explore longer-context video architectures, additional modalities, cached priors, distilled rewards, or online approximations to broaden CF-GRPO toward more comprehensive multimodal video reasoning.

\section{Conclusion}

We introduce Consensus Frame GRPO, a temporal-annotation-free process-level RL framework for Video-MLLMs. By constructing a consensus prior from intrinsic video cues and rewarding its agreement with a model-side frame-use score, CF-GRPO supplements outcome rewards with evidence-level feedback. Experiments show that VideoCFR improves several reasoning-oriented metrics and remains competitive on general video-understanding benchmarks, and ablations indicate that the consensus prior, salience-aware aggregation, and sharpening mechanism each contribute to the final performance. These results suggest that reward design for video reasoning should consider not only answer correctness, but also whether the generation process is aligned with consensus visual evidence.

\bibliographystyle{IEEEtran}
\bibliography{paper}

\end{document}